\documentclass[12pt]{article}
\usepackage{amsmath}
\usepackage{amsfonts}
\usepackage{graphicx,psfrag,epsf}
\usepackage{enumerate}
\usepackage{natbib}
\usepackage{apalike}
\usepackage{amsthm}
\usepackage{url} 
\newcommand{\blind}{1}

\expandafter\def\expandafter\normalsize\expandafter{%
    \normalsize
    \setlength\abovedisplayskip{7pt}
    \setlength\belowdisplayskip{7pt}
    \setlength\abovedisplayshortskip{7pt}
    \setlength\belowdisplayshortskip{7pt}
}

\usepackage[utf8]{inputenc}
\newcommand{\argmax}{\operatorname*{\arg\max}}
\newcommand{\argmin}{\operatorname*{\arg\min}}

\newcommand{\dd}[1]{\text{d}#1}

\usepackage[colorlinks=true,allcolors=black,breaklinks=true]{hyperref}
\usepackage{multicol}
\usepackage{multirow}
\usepackage{tabularx}
\usepackage{booktabs}
\usepackage{setspace}
\usepackage{algpseudocode}
\usepackage{hyperref}
\usepackage{subcaption}

\usepackage{titlesec} 
\titlelabel{\thetitle.\quad}

\usepackage{tikz}
\usetikzlibrary{arrows,automata,backgrounds,fit,patterns,petri,shadows,shapes}
\pgfdeclarepatternformonly{stripes}
{\pgfpointorigin}{\pgfpoint{0.25cm}{0.25cm}}
{\pgfpoint{0.25cm}{0.25cm}}
{
    \pgfpathmoveto{\pgfpoint{0cm}{0cm}}
    \pgfpathlineto{\pgfpoint{0.25cm}{0.25cm}}
    \pgfpathlineto{\pgfpoint{0.25cm}{0.12cm}}
    \pgfpathlineto{\pgfpoint{0.12cm}{0cm}}
    \pgfpathclose%
    \pgfusepath{fill}
    \pgfpathmoveto{\pgfpoint{0cm}{0.12cm}}
    \pgfpathlineto{\pgfpoint{0cm}{0.25cm}}
    \pgfpathlineto{\pgfpoint{0.12cm}{0.25cm}}
    \pgfpathclose%
    \pgfusepath{fill}
}
\tikzstyle{decisionA} = [regular polygon, regular polygon sides = 4, thick, minimum size = 1.25cm, inner sep = 0.1pt, draw = black, fill = gray!40]
\tikzstyle{decisionC} = [regular polygon, regular polygon sides = 4, thick, minimum size = 1.25cm, inner sep = 0.1pt, draw = black]
\tikzstyle{utilityA} = [regular polygon, regular polygon sides = 6, thick, minimum size = 1cm, inner sep = 0.1pt, draw = black, fill = gray!40]
\tikzstyle{utilityC} = [regular polygon, regular polygon sides = 6, thick, minimum size = 1cm, inner sep = 0.1pt, draw = black]
\tikzstyle{chanceA} = [circle, thick, minimum size = 1cm, inner sep = 0.1pt, draw = black, fill = gray!40]
\tikzstyle{chanceC} = [circle, thick, minimum size = 1cm, inner sep = 0.1pt, draw = black]
\tikzstyle{chance} = [circle, thick, minimum size = 1cm, inner sep = 0.1pt, draw = black, pattern = stripes, pattern color = gray!40]
\tikzstyle{empty} = [circle, line width = 0pt, minimum size = 1cm, inner sep = 0.1pt]

\newcommand{\Acal}{\mathcal{A}}
\newcommand{\Dcal}{\mathcal{D}}

\newcommand{\Pbb}{\mathbb{P}}
\newcommand{\opt}{*}
\newcommand{\possessivecite}[1]{\citeauthor{#1}'s \citeyearpar{#1}}

\begin{document}


\def\spacingset#1{\renewcommand{\baselinestretch}%
{#1}\small\normalsize} \spacingset{1}

\if1\blind
{
  \title{ \textbf{Adversarial Machine Learning: \\
    Bayesian Perspectives} \\ \small{(Published in the Journal of the American Statistical Association)} } 
  \author{David Rios Insua\footnotemark[1] (ICMAT-CSIC),\\ Roi Naveiro\footnote{These authors contributed equally. } (CUNEF Universidad),\\ Víctor Gallego\footnotemark[1] (ICMAT-CSIC and Komorebi AI Technologies)\\
    and \\
    Jason Poulos (Harvard Medical School)}
    \date{\vspace{-5ex}}
  \maketitle
} \fi

\if0\blind
{
  \bigskip
  \bigskip
  \bigskip
  \begin{center}
    {\LARGE\bf Adversarial Machine Learning: \\
    Bayesian Perspectives}
\end{center}
  \medskip
} \fi

\bigskip

\begin{abstract}
Adversarial Machine Learning (AML) is emerging as a major field aimed at 
protecting machine learning (ML) systems against security threats:
in certain scenarios there may be  
adversaries that actively manipulate input data to fool learning systems.
This creates a new class of security vulnerabilities that ML systems may face, and a new desirable property called adversarial robustness essential 
to trust operations based on ML outputs. 
Most work in AML is built upon a game-theoretic modeling of the conflict between a learning system and an adversary, ready to manipulate input data. This assumes that each agent knows their opponent’s interests and uncertainty judgments, facilitating inferences based on Nash equilibria.
However, such common knowledge assumption is not realistic in the security scenarios typical of AML. \textcolor{black}{  After reviewing such game-theoretic approaches}, we discuss the benefits that Bayesian perspectives \textcolor{black}{provide} when defending \textcolor{black}{ML-based} systems. 
We demonstrate how the Bayesian approach allows us to explicitly model our uncertainty about the opponent’s \textcolor{black}{beliefs and interests}, relaxing
unrealistic assumptions, and providing more robust inferences. We illustrate this approach in supervised learning settings, and identify relevant future research problems. 
\end{abstract}

\noindent%
{\it Keywords:}  
Bayesian Methods, Adversarial Risk Analysis, Security, Cybersecurity, Machine Learning.
\spacingset{1.44} 
\pagenumbering{arabic} 

\section{Introduction}
\label{sec:introduction}

Over the last decade, an increasing number of processes \textcolor{black}{have been} automated through 
machine learning (ML) algorithms, making it more crucial that these algorithms become robust and reliable if we are to trust operations based on their output. State-of-the-art
ML \textcolor{black}{methods} perform extraordinarily well on standard data,
but have been 
shown to be vulnerable to adversarial examples \citep{goodfellow2014explaining}, data instances targeted at
fooling them.
The presence of adversaries has
been highlighted in areas such as spam detection \citep{zeager2017adversarial}, 
computer vision \citep{goodfellow2014explaining}
and automated driving systems \textcolor{black}{(ADS, \citealp{caballero2021decision})}.  
In those contexts, algorithms should acknowledge the presence of possible adversaries to protect from their eventual data manipulations.
\cite{COMITER2019} provides a review from a policy perspective showing how many AI systems, including content filters, \textcolor{black}{and} military and law enforcement systems,
are vulnerable to attacks. As a motivating example, consider fraud detection: 
as ML algorithms are incorporated to such a task, fraudsters learn how to evade them. For instance, they could find out that making a huge transaction increases the probability of being detected, and instead would issue smaller transactions. 

As a fundamental assumption, ML systems rely on using 
independent and identically distributed (iid) data for both training and operations \citep{zhang2021adversarial}. However, the security aspects of ML, part of the emerging field of \textcolor{black}{Adversarial Machine Learning (AML)},
challenge such hypothesis, given the
presence of adaptive adversaries ready to intervene to modify the data and obtain a benefit. 
Stemming from the pioneering work in adversarial classification 
\citep{adversarialClassification2004}, the prevailing paradigm in AML has modeled the confrontation between learning-based systems and adversaries through game theory \citep{menache2011network}. This entails common knowledge (CK) assumptions
\citep{hargreaves2004game}, which are questionable in the security domain as adversaries try to hide and conceal information.
Thus, there is a need for developing a better 
founded paradigm:
as \cite{fan2019selective} point out, a framework that guarantees robustness of ML against adversarial manipulations in a principled manner is required. 

After providing an overview of key concepts and methods in AML emphasizing the underlying game theoretic assumptions, we suggest an alternative formal Bayesian decision theoretic \textcolor{black}{ framework based on Adversarial Risk Analysis (ARA, \citealp{adversarialRiskAnalysis2009}) and illustrate it in supervised learning settings.} We end by suggesting a research agenda.
 
\section{Motivating examples} \label{eg}
Two examples serve us to motivate key
issues in AML. They showcase how the performance of ML systems may considerably degrade under subtle data manipulations, suggesting the need to take into account the presence of adversaries.
\vspace{0.1in}

\noindent 
{\bf Case 1. Attacking spam detection algorithms.} \label{sec:ex2}
Consider spam detection, an example of content filters
which are at the backbone of many security systems.
We study the performance degradation of different algorithms under the Good-Words-Insertion attacks described in \cite{naveiro2019adversarial}: the adversary attacks spam emails by inserting at most two \textit{good words}\footnote{\textcolor{black}{Words that are common in legitimate email but rare in spam.}} into them. 
Table \ref{tab:cleanVSattack} presents the accuracy
of four standard \textcolor{black}{ algorithms --- {\em  naive Bayes, logistic regression, neural network (NN)} and {\em  random forests} --- when facing clean
and attacked data,\footnote{The logistic regression is applied with L1 regularisation, which is equivalent to performing maximum a posteriori estimation in a logistic regression model with a Laplace prior \citep{park2008bayesian}. 
The NN has two hidden layers.} using the Spambase Data Set from the UCI ML repository \citep{UCI}.} 
\textcolor{black}{ Accuracy means  and standard deviations 
are estimated via repeated hold-out validation over 10 repetitions \citep{kim2009estimating}.}
%
\begingroup
\renewcommand{\arraystretch}{0.5}
\begin{table}[!h]
\caption{\textcolor{black}{ Accuracy (with precision) 
 	of four algorithms on clean data
 	(untainted); attacked data and unprotected; protected through ARA  
 	during operations; protected through ARA during training.}}
\centering
{\small
\begin{tabular}[t]{ccccc}
\toprule
\textbf{Algorithm} & \textbf{  Untainted} & \textbf{ Unprotected}& \textbf{ARA op.} & \textbf{ARA tr.}\\
\midrule
Naive Bayes & $ 0.882 \pm 0.004 $ & $ 0.754 \pm 0.027 $ & $ 0.939 \pm 0.006 $ & $---$\\
Logistic Regression & $ 0.932 \pm 0.004 $ & $ 0.673 \pm 0.005 $ & $ 0.898 \pm 0.008 $ & 
$ 0.946 \pm 0.003 $ \\
Neural Network & $ 0.904 \pm 0.029 $ & $ 0.607 \pm 0.009 $ & $ 0.882 \pm 0.025 $ &
$0.960 \pm 0.002$ \\
Random Forest & $ 0.912 \pm 0.005 $ & $ 0.731 \pm 0.008 $ & $ 0.807 \pm 0.007 $ & $---$ \\
\bottomrule
\end{tabular}
}
\label{tab:cleanVSattack}%
\end{table}
\endgroup
%
%
Observe, columns 2 and 3,
the important loss in accuracy of the four algorithms: 
\textcolor{black}{ major performance degradation 
may affect them when ignoring the possible presence
of adversaries. Columns 4 and 5 are discussed in Section 5.}
\hfill $\triangle$
\vspace{0.1in}

\begin{figure}[h!]
\centering
\begin{subfigure}{.4\textwidth}
  \centering
  \includegraphics[width=.5\linewidth]{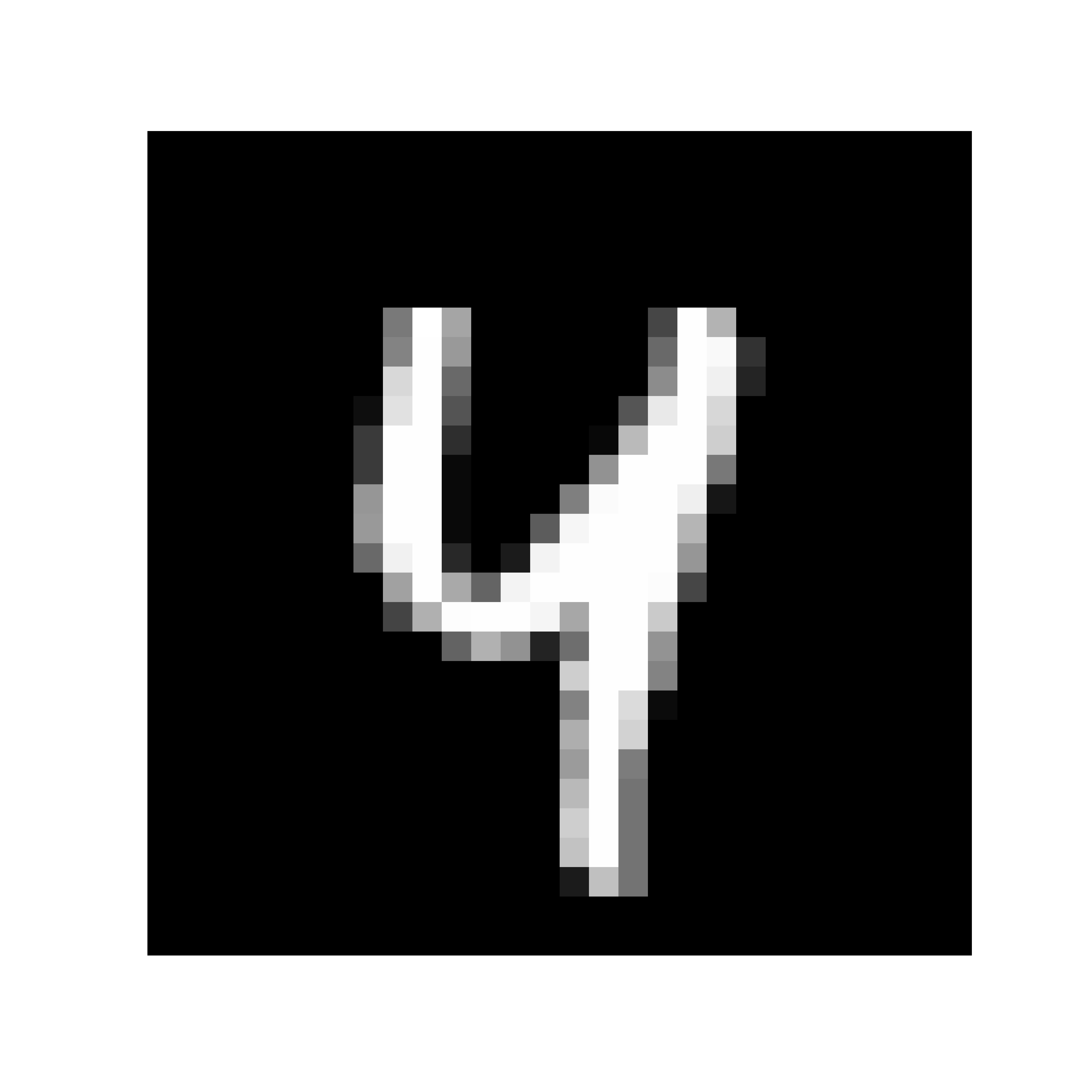}
  \caption{Original image.}
  \label{fig:2}
\end{subfigure}%
\begin{subfigure}{.4\textwidth}
  \centering
  \includegraphics[width=.5\linewidth]{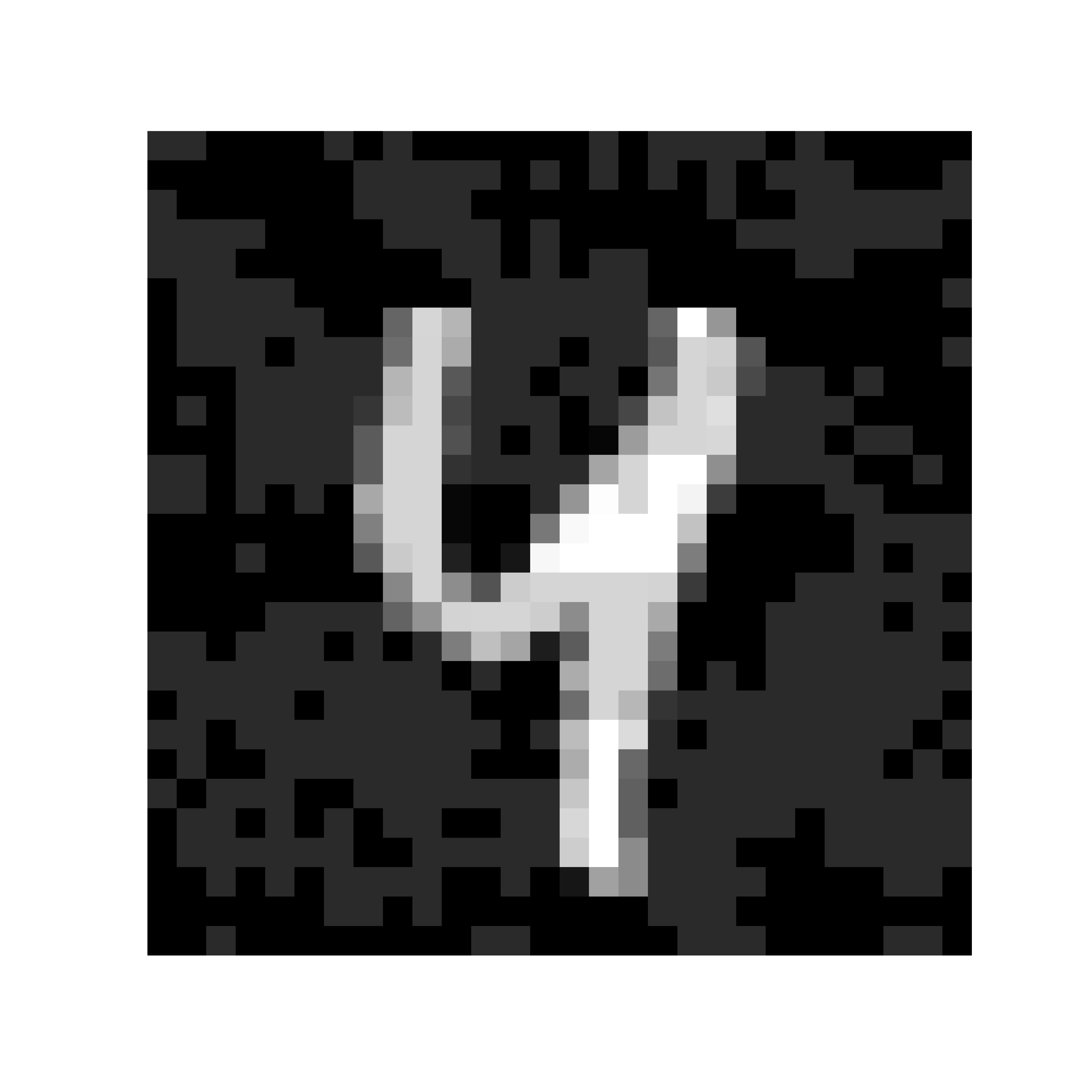}
  \caption{Attacked image.}
  \label{fig:7}
\end{subfigure}
\\
\caption{Original input and attacked version. }
\label{fig:49}
\end{figure}

\noindent{\bf Case 2. Attacking vision algorithms.}
 \textcolor{black}{Computer vision} algorithms are at the core of many AI 
applications such as  perception 
systems in ADS \citep{caballero2021decision}. 
The simplest and most notorious attack examples to
such algorithms  
consist of modifications of images so that the alteration
becomes irrelevant to the human eye, yet drives a model trained on millions of images to misclassify the attacked ones. This \textcolor{black}{attack} entails potentially relevant security consequences.
As an example, with a relatively simple convolutional NN \textcolor{black}{(CNN, }\citealp{DEEPLEARNING}), we achieve 99\% \textcolor{black}{test set} accuracy predicting the handwritten digits in the MNIST data set \citep{mnist}.
However, accuracy reduces to 59\% if we attack those data with the
{\em fast gradient sign method} \textcolor{black}{(FGSM, }  \citealp{szegedy2013intriguin}).  Fig.~\ref{fig:49} provides examples of 
an original MNIST image and an attacked one. Our CNN correctly classifies the original image (left) as a 4; however, it misclassifies the attacked one (right) as a 9.
%
FGSM and related attacks described below are easily built through low cost computational methods. In certain settings, they require the attacker to have precise knowledge about the architecture of the corresponding predictive model.
This is debatable in most security settings and is a driving force in this paper.
\hfill $\triangle$

\section{Adversarial Machine Learning: A Review}

We review \textcolor{black}{the main} results and concepts in AML.
We focus on key ideas in the brief history of the field, motivating a reflection that will lead to our alternative Bayesian framework in Section \ref{KK}. 
Further perspectives may be seen in \cite{vorobeychik2018adversarial}, \cite{joseph2019adversarial}, \cite{BIGGIO2018317} and \cite{dasgupta2019survey}.
We describe first the usual workflow in AML within which most previous research can be embedded.
In general, we  refer to the learning system as \textit{defender} ($D$, she), and to the adversary manipulating data as \textit{attacker} ($A$, he).

\subsection{An Adversarial Machine Learning workflow}\label{sec:wf}

Guaranteeing protection of a learning system against attacks involves undertaking various activities related to security evaluation, threat modeling, and attack simulation. 
Let us frame \textcolor{black}{these} activities within a workflow for AML with three steps, as in \cite{BIGGIO2018317}: \textcolor{black}{(1) gathering intelligence to study the likely attacks that a system may face; (2) forecasting likely attacks; and (3), protecting learning systems from such attacks.} 
\vspace{0.1in}

\noindent{\bf 1. Gathering intelligence.}
This activity is critical to ensure ML security in adversarial environments. Obviously, any algorithm could be fooled if adversarial data modifications are not somehow restricted: in an extreme case, if an instance in a binary classification problem is modified so that it is indistinguishable from an instance of the other class, clearly, the algorithm would misclassify such instance. However, the adversary is probably not interested in making such data modifications as the new instance might lose its malicious purposes. 
Thus, an in-depth study of likely attackers is key in AML. In general, we should gather information about three attacker features.

  
First, we assess his {\em goals}\textcolor{black}{, which may 
range from appropriating funds to causing harm to people or organizations} \citep{couce2019gira}. 
Prior to deploying a ML system, 
it is crucial to guarantee robustness against attackers with the most common goals.
For instance, in fraud detection, the attacker usually obfuscates fraudulent transactions to make the system classify them as legitimate in search of an economic benefit: a fraud detection system should be robust against such attacks, trying to minimize economic losses. In general, attacker goals are  classified 
along two dimensions. For the first one, {\em violation type},
a usual distinction is between \textcolor{black}{  {\em integrity} (aimed at moving the prediction about particular instances towards the attacker's target, e.g., to have  malicious samples misclassified as legitimate); {\em availability}
(aimed at increasing the predictive error to make the system unusable) and {\em privacy} (exploratory attacks to gather information about the ML system) {\em violations}. The second dimension refers to {\em attack specificity}, where the usual distinction is between {\em targeted attacks}, addressing a few specific defenders, and {\em indiscriminate attacks}, affecting many defenders in a random manner \citep{rios2019adversarial}.}
%


Second, we assess the {\em knowledge} that the adversary could have about 
the ML system. 
At one end of the spectrum, we find {\em white box} or {\em perfect knowledge} attacks: the adversary knows every aspect of the system. This is almost never the case in security scenarios, except perhaps for insider attacks \citep{joshi}. Yet, they could be useful in sequential settings where the ML system moves first, training an algorithm to fit its parameters; \textcolor{black}{and} the adversary, who moves afterwards, has time to observe the behavior of the system and learn about it.\footnote{However, although the adversary may have some knowledge,  assuming that this \textcolor{black}{knowledge} is perfect is not realistic and has been criticized \citep{adversarialClassification2004}.}
At the other end, {\em black box} or {\em zero knowledge} attacks assume that the adversary has capabilities to query the system but does not have any information about the data, feature space, or specific algorithms used. This is the most reasonable assumption when attacking and defending decisions are made simultaneously. In between, attacks are called {\em gray box} or {\em limited knowledge}, the most common type of attacks in security settings, especially when attacking and defending decisions are made sequentially but there is private information that agents are not willing to share. 

Finally, the third feature refers to the
adversary's {\em capabilities} 
to influence on data and other features. 
With {\em poisoning attacks}, he may obfuscate training data to later induce errors during operations. Alternatively, {\em evasion attacks} have no influence on training data, but perform modifications during operations, for instance when trying to evade a detection system.
These data crafting activities are typical in 
AML and designated to come from a so-called \emph{data-fiddler}. There could be as well attackers capable of changing the underlying structure of the problem affecting process parameters, called
\emph{structural attackers}. Moreover, some 
adversaries could be making  decisions 
in parallel to those of the defender with the agents' losses
depending on both decisions, which
we term \emph{parallel attackers}.\footnote{Some attackers could combine the three capabilities in certain scenarios.
For example, in cybersecurity an attacker
might add spam modifying its proportion (structural);
alter some spam messages (data-fiddler); and, in addition, undertake his own business decisions (parallel). See
\cite{gil}.}
\vspace{0.1in} 

\noindent 
{\bf 2. Forecasting likely attacks.}
In our path to enhance protection, once having gathered intelligence about the potential attacks to a learning system, we should produce models for how the adversary may behave when facing new data. A central argument for us
is that such models must take into account our uncertainty about adversarial aspects. As mentioned, most previous research along these lines has been usually based on game theory assuming full knowledge about the adversary. Thus, given some data, the adversary would behave deterministically:
the standard approach 
forecasts
attacks solving constrained optimization problems with different assumptions about the adversary’s knowledge, goals and capabilities.
The corresponding objective function 
assesses attack effectiveness, taking into account \textcolor{black}{the assumptions about the attacker features}.
The constraints frame assumptions such as the adversary wanting
to avoid detection 
or having available a maximum attacking budget.
%
%
However, full knowledge assumptions are generally unrealistic in the AML realm as adversaries try to conceal information: adversary modeling must take into account \textcolor{black}{the} lack of information and corresponding uncertainty that we have about the adversary. Beyond the {\em aleatoric} and {\em epistemic} uncertainties typical in risk analysis, in AML, 
analysts need to consider as well {\em concept uncertainty}
\citep{banks2020adversarial}. \textcolor{black}{Thus}, given some data, we associate an \textit{attacking model} with a probability distribution over attacks which encodes our uncertainty about how the adversary will act when seeing a particular instance.\footnote{Previous attacking models can be easily recovered in this framework assuming degenerate distributions as later illustrated.}





\vspace{0.1in}

\noindent{\bf 3. Protecting ML algorithms.} 
Once relevant adversarial models have been produced, the last step consists of protecting learning systems against the modeled attacks. 
Broadly speaking, two types of defenses have been proposed. \emph{Reactive defenses} aim to mitigate, even eliminate, the effects of an eventual attack. They include timely detection of attacks \citep[e.g.,][]{naveiro2019large}; frequent retraining of learning algorithms; or verification of algorithmic decisions by experts. The second type, \emph{proactive defenses},
aim to prevent attack execution. They can entail
{\em security-by-design} approaches such as explicitly accounting for adversarial manipulations \citep{naveiro2019adversarial} 
or producing provably secure 
algorithms against perturbations \citep{gowal2018effectiveness}; or {\em security-by-obscurity} techniques such as randomization of algorithm responses, or gradient obfuscation to make attacks less likely to succeed \citep{athalye2018obfuscated}.
A more interesting classification of defenses, later emphasized, refers to whether the protection is carried out at {\em training} or at {\em operation} time. Defenses of the former class, train learning systems robustly, anticipating future adversarial attacks. Defenses of the latter class, when receiving a potentially attacked instance, undertake inference about possible originating instances to make the corresponding decision.

\subsection{Core concepts in Adversarial Machine Learning}\label{sec:review}

Most work in AML has dealt with supervised learning facing adversarial threats, with a focus on either proposing new attacks to learning systems to showcase their vulnerabilities (thus, related to steps 1-2 of the above workflow) or proposing defenses to protect \textcolor{black}{algorithms} from common attacks (step 3). 
\vspace{0.1in} 

\noindent{\bf Attacks to learning systems.}
The most common goal of proposed attacks is to modify instance covariates to induce the learning system into making wrong decisions upon observing or analysing such contaminated covariates.
One of the most influential concepts triggering the current interest in AML are {\em adversarial examples}. They are introduced within NN models as perturbed data instances aimed at fooling NNs, obtained
through solving certain optimization problems \citep{szegedy2013intriguin}. These models are highly sensitive to such examples; recall case 2 in Section \ref{eg}. 

Adversarial examples have traditionally targeted computer vision systems, using techniques such as FGSM 
or \emph{projected gradient descent} (PGD, \citealp{madry2018towards}). These techniques find a constrained perturbation of an image that maximizes the loss function used to train the computer vision system. This optimization is usually approximated using gradient ascent routines.
In addition, attacks like these have been extended to target other systems such as natural language processing \citep{zhang2020adversarial},
due to their increasing relevance. 
Attacking strategies targeting tabular data are usually application specific. Most approaches model the confrontation between the attacker and the learning system as a game \citep{bruckner2012static}. Assuming that each agent knows their opponent’s interests and uncertainty judgments, the adversary will perform the attacks dictated by the Nash equilibrium strategy of that game or emerging as best responses.

Much less AML work is available in relation with unsupervised learning. \cite{kos2018adversarial} describe adversarial attacks to generative models,  
where slight perturbations to the model input may yield a reconstructed output that is very different from the original input.
\cite{biggio2013data} study clustering under adversarial disturbances. The authors describe how to create attacks that significantly alter cluster assignments, as well as obfuscation attacks that slightly perturb inputs to be clustered in a predefined assignment, showing that single-link hierarchical clustering is sensitive to such attacks. 
Lastly, adversarial attacks targeting time series forecasting systems have started to attract interest. \cite{alfeld2016data} describes an attacker manipulating the inputs to drive the latent space of a linear autoregressive (AR) model towards a region of interest. Similarly, \cite{papernot2016crafting} propose adversarial perturbations over recurrent NNs, and \cite{naveiro2021adversarial} studies adversarial attacks against Bayesian dynamic models.
\vspace{0.1in}

\noindent{\bf Defenses against attacks.} 
As mentioned in Section \ref{sec:wf}, we distinguish two types of AML defenses: those that promote protection strategies during training, and those that protect during operations. \textcolor{black}{All these defenses assume that a clean training set is available and the attacks happen once the ML system is deployed, the most common case in realistic scenarios.}

\subparagraph{Protection during operations.}

The pioneering defense of this type, proposed in \cite{adversarialClassification2004}, was devoted to protect classification systems. Given the importance of classification in many cutting-edge ML applications, this paper opens up a research area known as  \textit{adversarial classification} (AC). 
The authors 
view AC as a game between a classifier (defender), and an adversary (attacker). During operations, upon observing a new vector of covariates, 
$D$ aims at finding an optimal classification strategy against $A$'s optimal attacks. Computing Nash equilibria  
in such general games quickly becomes very complex. Thus, the authors propose a forward myopic version: \textit{D} first assumes that data is untainted, computing her optimal classification decision; then, \textit{A} deploys his optimal attack against it. Subsequently, \textit{D} implements her best response against such attack, and so on.
They assume CK as all parameters of both players are known to each other. \textcolor{black}{ Although standard in game theory, this assumption is unrealistic in security settings typical in AML, an issue acknowledged in \cite{adversarialClassification2004} and largely unsolved.}

Subsequent AC approaches, reviewed in \cite{biggio2014security}, have focused on analyzing attacks over algorithms and upgrading their robustness against them, always making strong assumptions about the adversary.
For instance, \cite{adversarialLearning2005} consider that 
the attacker can send membership queries to the classifier to issue optimal attacks.
Other approaches have focused on improving Dalvi et al.'s model but, as far as we know, none have disposed of the unrealistic CK assumptions.



\subparagraph{Protection during training.}

An important family of AML defenses try to robustify learning systems by modifying the way training is performed.
A relevant source of AML defenses of this type are {\em Adversarial Prediction Problems} (APPs), whose focus is 
on building adversarially robust predictive models. It is assumed that during operations an adversary that exercises 
some control over the data generation process will be present: 
the data generation distributions at operations and training will be different, jeopardizing standard prediction techniques.
%
To address this, APPs
 model interactions between the predictor and a fictitious adversary during training as a 
two-agent game with a system aimed at learning a parametric predictive model 
and an adversary trying to transform the distribution
governing data. 
The predictor will therefore minimize the expected cost under the operations data distribution. 
In turn, the fictitious adversary will modify data optimizing his expected cost under this distribution. As such distributions are not known, agents optimise their regularized empirical costs, based on training data: 
the predictor chooses her model minimizing her expected cost with respect to an attacked version of the training data chosen so as to minimize the adversary's cost. The final optimization problems are case dependent.


In {\em Stackelberg prediction games}, \cite{bruckner2011stackelberg} assume full information
of the attacker about the predictive model used by the defender who, in addition, has full information about the adversary's costs and action space. $D$ acts first choosing her parameters; then, $A$, who observes this decision, chooses the optimal data transformation. 
Finding NE in these  games leads to a bi-level optimization problem, minimizing the defender's cost function subject to 
the adversary, after observing the defender's choice, minimizing his cost function. As nested optimization problems are intrinsically hard, the authors restrict to simple classes where analytical solutions are
available.\footnote{More recently, \cite{naveiro2019gradient} provide efficient gradient methods to approximate solutions in more general problems.}
%

In {\em Nash prediction games} both agents act simultaneously. 
\citet{bruckner2012static}  provide conditions for existence and uniqueness of NE in certain subclasses of these games.
Notice that APPs propose training using instances that are attacked by ``fictitious attackers,'' and hope that this will serve as a proxy for dealing with real attackers. However, it \textcolor{black}{is} assumed that the fictitious attackers' costs and probabilities are CK, which is not realistic
in security scenarios. \textcolor{black}{ Deviations from the assumed attackers' models potentially lead to severe performance degradation as we later illustrate.}

Another important family of defenses that affect the training stage are those that aim at robustifying models against  adversarial examples.
\textcolor{black}{  {\em Adversarial training} (AT) \citep{madry2018towards} is the most important one.} It  aims at choosing a parametric model (usually a NN) that minimizes the empirical risk evaluated under worst-case data perturbations. Thus, it can be viewed as a zero-sum version of an APP: in AT, the fictitious attacker is assumed to select the data manipulation that maximizes the defender's costs within some constrained region. AT approximates the inner optimization through the PGD algorithm, ensuring that the perturbed input falls within a tolerable boundary, usually specified through some restriction on a norm distance. 
  Attack complexity depends on the chosen norm.
However, recent pointers urge modellers to depart from using 
norm based approaches \citep{carlini2019evaluating} and develop more realistic attack models as in \possessivecite{brown2017adversarial} adversarial patches.

\cite{liu2018adv} adapted the idea of AT to 
Bayesian NNs, \textcolor{black}{ using the notion that incorporating randomness in the NN weights enhances their robustness}. They propose training Bayesian NNs using mean-field variational inference. However, as in AT, instead of maximizing the evidence lower bound (ELBO) under the original training instances, they propose maximizing the ELBO under a worst-case attacker that chooses the best data manipulation inside a ball in a normed space.
As with AT, this heuristic implicitly assumes full knowledge in the construction of the fictitious attacker, not taking into account the existing uncertainty. This may produce performance degradation when dealing with actual, unknown attackers. Finally, another relevant but more heuristic family of defenses is called \textit{adversarial logit pairing} \citep{kannan2018adversarial} in which logits of pairs of attacked and clean instances are encouraged to be the same, thus yielding the same prediction for both.

\textcolor{black}{All defenses presented assume full knowledge in the attacking models,  leading to deterministic attacks. Taking into account existing uncertainties about adversaries would be crucial to produce sensible defenses. This is the goal of the Bayesian framework for AML presented in Section \ref{sec:wf}.}

\vspace{0.1in} 

\noindent{\bf Adversarial Reinforcenment Learning.}
While there is considerable AML research in supervised and unsupervised learning, much less work is available in relation to reinforcement learning (RL). In it, adversarial aspects refer to the presence of agents whose decisions affect the reward perceived by our supported agent.
The prevailing solution approach in standard RL is $Q$-learning \citep{sutton1998reinforcement}; its 
adaptation to large problems, deep Q-learning, has faced an incredible growth recently \citep{silver2017mastering}. It relies on models, such as convolutional NNs, to process input information. \textcolor{black}{    Consequently, the 
adversarial examples described above apply when fooling RL systems \citep{10.5555/3172077.3172414}.}

Single-agent RL methods fail in presence of other agents that interfere with their learning process, as they do not take into account the non-stationarity due to the other agents' actions: $Q$-learning may lead to sub-optimal results \citep{marl_over}.  Thus, a deployed RL system must be able to reason about and forecast the adversaries' behavior. Several methods to enhance $Q$-learning in multiagent systems have been proposed, mostly focusing on adapting ideas from game theory into RL, 
%
mainly focusing on modeling the multiagent system through Markov games. 
Three well-known solutions \citep{tuyls2012multiagent} are minimax-$Q$ learning; 
Nash-$Q$ learning; 
and friend-or-foe-$Q$ learning, but these come with unrealistic CK assumptions or can only be applied in restrictive scenarios.

\vspace{0.1in}

\noindent{\bf Further comments.}
Practically all ML methods have been touched upon from an adversarial perspective. 
Of major importance in this field is the 
\texttt{cleverhans} \citep{papernot2018cleverhans} library, aimed at accelerating research in developing new attack threats and more robust defenses specifically for deep neural models.

AML is a  difficult area which evolves rapidly and leads to an arms race in which the community alternates cycles of proposing attacks and implementing defenses that deal with them. Thus, it is important to develop sound techniques. Note that, stemming from \cite{adversarialClassification2004}, most of AML research 
has been framed, \textcolor{black}{ sometimes implicitly,}  within a standard game theory approach characterized by NE and refinements. However, these entail CK assumptions 
which are hard to maintain in the security contexts typical of AML. 
We next propose a Bayesian decision theoretic methodology to solve AML problems, 
utilizing an ARA perspective \citep{adversarialRiskAnalysis2009} to model the confrontation between attackers and defenders mitigating questionable CK assumptions. 




\section{A Bayesian Workflow for AML}\label{KK}

We now revisit the workflow in Section \ref{sec:wf} proposing a Bayesian decision theoretic alternative to AML. It is based on ARA,
which operationalizes the Bayesian approach to games \citep{kadane1982subjective} and facilitates a procedure to forecast adversarial attacks. 
%
%
\textcolor{black}{ ARA provides prescriptive support to a decision maker (DM), the ML system in our case, facing one or more 
attackers whose actions affect her decision making process. The DM is assumed to be a rational, expected utility maximizing agent. Her utility 
and beliefs (epistemic uncertainty) depends on her decision, the adversaries' rationality and decisions (concept uncertainty), and possibly some other random variables (aleatoric uncertainty).} Since CK is not assumed, random variables model adversaries' decisions that must be integrated out to compute  expected utilities. ARA provides a coherent procedure to obtain probabilistic forecasts of adversaries' actions. The main idea is to model the adversaries' decision making process, putting priors on unknown quantities to reflect the lack of knowledge. This way, the optimal adversarial decision becomes probabilistic. Simulations from such random optimal decisions are used to compute the DM's expected utility. 

Our focus will be on protecting 
a supervised learning system 
($D$) which receives instances described by covariates $x \in \mathbb{R}^d$, with each instance having an associated output $y$.
Uncertainty about the instances' output given its covariates is modeled through a distribution $ p(y| x)$.  This distribution can arise from a generative model, where distributions $p(x)$ and $p(x|y)$ are modeled  explicitly and $ p(y| x)$ is obtained via Bayes formula; or from a discriminative model, in which $ p(y| x)$ is modeled directly  \citep{bishop}. 
It may be derived through maximum likelihood or in a Bayesian way using training data which, by assumption, is free of attacks.
Whichever estimation method is adopted, upon observing a new instance with covariates $x$, the Defender must decide the corresponding output. As $D$ is rational, 
she decides based on maximum \textcolor{black}{predictive} utility through
$$
\argmax_{y_D} \int u (y_D , y ) p (y | x ) \dd y,
$$
where $u (y_D, y)$ is the utility that she perceives when an instance whose actual output is $y$ is assigned output $y_D$.\footnote{If the output is discrete rather (as in  classification problems) the integral is replaced by a sum.} In adversarial settings, \textcolor{black}{ agent $A$ applies an attack $a$ to the features $x$} leading to the transformation $x'=a(x)$, the observation actually received by $D$. 
We focus on exploratory attacks, \textcolor{black}{ which affect just over operational data, leaving training data untainted. 
Let us revisit the three stages of the workflow.} 
\vspace{0.1in}


\noindent {\bf 1.  Gathering intelligence.} 
This stage entails modeling the attacker's 
problem. Assessing his goals requires determining the actions that he may undertake and the utility that he perceives when performing a specific action, given a defender's strategy: the
output is the set of attacker's decisions and a functional form for his utility,  generally dependent on his and the defender's decisions. Assessing the attacker knowledge entails looking for information that he may have when performing the attack, and his degree of knowledge about it, as we do not assume CK. This
requires not only a modeling activity, but also a {\em security 
assessment} of the ML system determining which of its elements \textcolor{black}{(training data, feature space, architecture, loss function, parameters, etc.)}
are accessible to the attacker. 
Finally, identifying his 
capabilities requires determining which part of the defender
problem the attacker has influence on. 

Consider an adversary aiming at fooling a supervised learning system by modifying the value of the covariates. The adversary receives objects with covariates $x$ and output $y$ and
manipulates $x$, transforming them into $x'$. His goal is to induce the defender to make non-optimal decisions for the output corresponding to the observed covariates.
Following a normative decision theoretic perspective, we model the adversary as a rational agent choosing data manipulations to maximize expected utility. Let $u_A(y_D, y)$ be the adversary's utility when the defender assigns output $y_D$ to an instance whose actual output is $y$. This utility can also depend on the specific data manipulation, as distinct manipulations can entail different costs; however,
to simplify notation we do not include explicitly this dependence.
The adversary thus chooses data manipulations through
\begin{equation} \label{adv}
    x' (x, y) = \argmax_{z} \int u_A(y_D, y) p_A(y_D \vert z = a(x)) \dd y_D,
\end{equation}
where $p_A(y_D \vert z = a(x))$ models the adversary's belief about the defender's decision upon observing the manipulated instance $z = a(x)$. 
\vspace{0.1in}



\noindent
{\bf 2. Forecasting likely attacks.} 
Based on step 1, we produce models for how the adversary \textcolor{black}{ would modify data, encoding not only the information gathered}, but also our uncertainty about the adversary's elements. The output of this stage is an \textit{attacking model}, a probability distribution over adversarial manipulations that encodes all relevant uncertainties.
A formal Bayesian way to do this, as suggested by ARA, is to place priors on every unknown element of the adversary's decision making problem. The uncertainty implied by these priors is propagated to the optimal adversarial data modification that becomes random. Its associated probability distribution conforms to the attacking model used to protect the learning algorithm.
In general, evaluating analytically such model will be unfeasible. However,
\textcolor{black}{ in most cases it is conceptually and computationally
simple to sample from it. This  just entails sampling from our priors and, for each sample, solving the adversary's decision making problem,
which provides} a sample from the random optimal data manipulation.
In our adversarial supervised learning context,
the adversary 
will produce data manipulations solving \eqref{adv}. Under standard CK assumptions, our \textit{attacking model} would be
%
  $  p(x' \vert x) = \delta \left( x' - \argmax_{z} \int u_A(y_D, y) p_A(y_D \vert z ) \dd y_D \right).$
%
However, \textcolor{black}{ as argued, CK rarely holds in security domains there being  } multiple sources of uncertainty. First of all, unlike the adversary, we do not know the actual output $y$ for a given instance $x$. Thus, our \textit{attacking model}  must account for this uncertainty through $p(x' \vert x) = \int p(x' \vert x, y) p(y \vert x) \dd y$.
Sampling from $p(y \vert x)$ is standard. 
Sampling from $p(x' \vert x, y)$ is more complex, as we usually have uncertainty about the adversary's utility $u_A(y_D, y)$ and his probability estimates $p_A(y_D \vert z )$. We propose modeling such uncertainty with, respectively, random utilities $U_A$ and random probabilities $P_A^{y_D}$ defined, without loss of generality, over an appropriate common probability space 
$(\Omega,{\cal A},{\cal P})$ with atomic elements $\omega \in \Omega$. 
This induces a distribution over the Attacker's optimal attack defined through
\begin{equation*}
    X_{\omega }' (x, y) = \argmax_z \int U^\omega_A(y_D, y) P^\omega_A(y_D \vert z ) \dd y_D ,  
\end{equation*}
 \textcolor{black}{ leading to } $p(x'|x, y)= {\cal P} (X_{\omega }' (x, y)= x')$. In such a way, our model for $p(x'|x, y)$ properly accounts for the existing uncertainty about the adversary.
By construction, if we sample utilities and probabilities from their corresponding priors and solve \eqref{adv}, this solution would be distributed according to $p(x' | x, y)$. Overall, to produce samples from $p(x' \vert x)$, we first sample from $y$ from the posterior predictive distribution $p(y | x)$, and then $x'$ from $p(x' | x, y)$.

The specifications of random utilities and random probabilities are application-specific. Guidelines for adversarial classification are given in \cite{gallego2020protecting}. Notice that, to model our uncertainty about the adversary, we study his decision making problem from our point of view. Obviously, when analyzing the Attacker's problem, we must take into account his uncertainty about our elements; e.g., his uncertainty about the defender's decision upon observing the manipulated instance. This could lead to a infinite hierarchy of decision making problems as presented in \cite{rios2012adversarial}, albeit in a simpler context. 
One would typically model several steps in the hierarchy and stop at a level in which no more information is available. At that stage, non-informative priors over the involved probabilities and utilities can be used.

To sum up, we have now a general, formal, decision-theoretic alternative to produce attacking models, that is, samples from $p(x' \vert x)$, keeping CK assumptions at a minimum. Note, however, that previous attacks proposed in the literature can be adopted within the proposed workflow. For instance, consider the FGSM attack (case 2, Section \ref{eg}): it assumes
that the defender uses a parameterized model with parameters $\theta$,
 trained minimizing a  loss function $L(\theta, x, y)$;
the attacker has full knowledge about such loss, or at least its gradient
and has resources to perturb the covariates by adding a small vector $\epsilon$. Under this CK setting, the proposed (deterministic) attack is $x' = x + \epsilon \cdot \text{sign} \left[ \nabla_x L(\theta, x, y) \right]$, leading to an attacking model $p(x' \vert x, y, \theta)$,
degenerated at such $x'$. 


\textcolor{black}{Our proposed attacking model would be classified as gray box since, even if perfect knowledge is not assumed, certain assumptions about the adversary are being made such as him being an expected utility maximizer. One of the advantages of the proposed approach is its ability to create more complex attacking models; e.g., through mixtures of attackers with different solution concepts \citep{risa.12439}. 
Finally, for purely black-box settings in which the attacker is only assumed to have query access to the target learning system, an interesting approach to produce attacking models has been recently introduced \citep{lee2022query}. Here, the authors propose to generate attacks using Bayesian optimization, modelling the (unknown) attacker's objective function with a Gaussian process that is sequentially updated after the queries' results are received.}

\vspace{0.1in}
\noindent 
{\bf 3. Protecting ML algorithms.} 
Once with a reasonable probabilistic \textit{attacking model}, we  protect our learning system against such attacks,
either at operations or training. 
\subparagraph{Protection during operations.} 
The Defender observes a potentially attacked vector of covariates $x'$.
Based on it, she has to assign an output $y_D$. An adversary-unaware $D$ will make this decision by maximizing the posterior predictive utility, $\argmax_{y_D} \int u(y_D, y) p(y \vert x') \dd y.$  \textcolor{black}{In terms of the spam detection example (Section \ref{eg}), $x'$ would represent the words used in an email, potentially manipulated by a spammer, and $y_D$ corresponds to labeling such email as spam or legitimate. 
As illustrated, solving this classification using an adversary-unaware $D$ could lead to serious performance degradation. }

Upon observing $x'$, $D$ is uncertain about the actual originating vector of covariates $x$ \textcolor{black}{(the words of the originating email)}. She may model this uncertainty through a distribution $p(x \vert x')$ and decide based on the posterior predictive utility, where $x$ has been marginalized out
\begin{equation}\label{def}
\argmax_{y_D} \int u(y_D, y) \left [\int  p(y \vert x) p(x \vert x') \dd x \right] \dd y,
\end{equation}
where conditional independence of $y$ and $x'$ given $x$ is assumed. Thus, 
when adversaries are present, rather than deciding based on the posterior predictive distribution of a new instance, we do it based on what we designate the {\em robust adversarial posterior predictive distribution} (RAPPD) $\int  p(y \vert x) p(x \vert x') \dd x$.
This is generally not available in closed form and has to be evaluated numerically using Monte Carlo methods.
The key step for this is the ability to sample from $p(x \vert x')$, that is, the distribution of possible originating covariates given the observed ones $x'$. 
Here is where the attacker models from Step 2 come into play. Having constructed a model for $p(x' \vert x)$ and being able to sample from it, all we need is to generate samples from the inverse distribution $p(x \vert x')$. Techniques for this based on Approximate Bayesian Computation (ABC) are discussed in \cite{gallego2020protecting}.
\textcolor{black}{In terms of the spam detection case, the attack model $p(x' \vert x)$ would reflect $D$'s uncertainty about the manipulated words $x'$ that the adversary selects, given the email with words $x$.}

\subparagraph{Protection during training.} 
As mentioned, other defenses modify how training is performed in order to take into account the possible presence of an adversary during operations. Their goal is to train using artificial data that somehow mimic actual, potentially attacked, operational data through several heuristics. 
Most of them model how the attacker would modify the instances in the training set. 
Having trained the classifier in this manner, 
$p (y | x')$ could be directly evaluated 
at the operation stage as this probability has been inferred taking into account the presence of an attacker.
As discussed, these methods assume models for how the attacker would modify training instances that do not take into account the existing uncertainty. 
For instance, AT, as a proxy to robustify classifiers against attacks, considers an attacker that produces the worst data modification for the classifier,
assuming explicitly that the classifier has knowledge about the attacker's objectives, and implicitly that he has knowledge about the classifier's utility.
However, in realistic settings, we would not have precise information about how the attacker modifies a given instance, as we do not know, in general, his intentions and probability assessments. Thus, assuming a deterministic attack for robustification purposes may result inappropriate. We believe that it is crucial to account for such uncertainty explicitly. 

\cite{ye2018bayesian} take a step in this direction. They provide a Bayesian counterpart of AT, designated Bayesian Adversarial Learning,
assuming that the Defender has observed clean training data $\mathcal{D} = \lbrace x_i, y_i \rbrace_{i=1}^N$ which are samples from an unknown distribution. Based on this, an adversary unaware defender using a model parameterized by $\theta$ will simply compute the posterior $p(\theta \vert \mathcal{D})$ and employ this to calculate the predictive distribution used during operations. However, the presence of an adversary at operations changes the data generation mechanism, Thus, using the original $\mathcal{D}$ for inference could lead to large performance degradation. Instead, the authors suggest computing a {\em robust adversarial posterior distribution} (RAPD) over the parameters
%
$\int  p(\theta \vert \tilde{ \mathcal{D} }) p(\tilde{ \mathcal{D} } \vert \mathcal{D} )  \dd \tilde{\mathcal{D}}$, where $\tilde{ \mathcal{D} }$ refers to the manipulated training data.
%
Gibbs sampling provides samples from it iterating through
\begin{eqnarray}
\label{gibbs1}
\tilde{\mathcal{D}}^{(t)} \vert \theta^{(t-1)}, \mathcal{D} &\sim& p(\tilde{\mathcal{D}} \vert \theta^{(t-1)}, \mathcal{D}), \\
\label{gibbs2}
\theta^{(t)} \vert \tilde{\mathcal{D}}^{(t)} &\sim & p(\theta \vert \tilde{\mathcal{D}}^{(t)}).
\end{eqnarray}
After a burn-in period, samples $\lbrace \theta^{(T)} , \tilde{\mathcal{D}}^{(T)}\rbrace$ follow the joint posterior $p(\theta, \tilde{\mathcal{D}} \vert \mathcal{D})$ and, consequently, sample $\theta^{(T)}$ follows the RAPD.
The distribution $p(\tilde{\mathcal{D}} \vert \theta, \mathcal{D})$ quantifies our uncertainty about the data generation process; i.e., about how the adversary will modify data $\mathcal{D}$. 

As with attacks, if we assume a high degree of CK, earlier defense mechanisms can be framed within our workflow. In AT, the defender uses a parametric model with parameters $\theta$. An adversary unaware $D$ makes inference about $\theta$ minimizing a loss function $\sum_{i=1}^N L(\theta, x_i, y_i)$, with $N$ training points. When taking into account the adversary, AT proposes minimizing
$
\sum_{i=1}^N \max_{\Vert \gamma \Vert \leq \epsilon} L(\theta, x_i + \gamma, y_i);
$
that is, minimize the loss evaluated under worst case perturbations in some constrained region. 

Most common losses can be written as negative log posterior distributions.
Thus, for the rest of the discussion, assume that the loss function can be written as
\begin{equation} \label{aux}
    \sum_{i=1}^N L(\theta, x_i, y_i) = - \sum_{i=1}^N \log p(x_i, y_i \vert \theta ) - \log p(\theta).
\end{equation}
%
%
If we assume that, for some fixed value of $\theta$, $p(\tilde{\mathcal{D}} \vert \mathcal{D})$ has the form
\begin{eqnarray*}
p(\tilde{\mathcal{D}} \vert \mathcal{D}) = \prod_{i=1}^N p(\tilde{x}_i, y_i \vert x_i, y_i) = \prod_{i=1}^N \delta \left(\tilde{x}_i - \left[  x_i + \argmax_{\Vert \gamma \Vert \leq \epsilon} L(\theta, x_i + \gamma, y_i)\right] \right),
\end{eqnarray*}
and we recover AT as a maximum \emph{a posteriori} (MAP) estimate of $\theta$ under the robust adversarial posterior distribution. Indeed, notice that the robust posterior can be written
\begin{eqnarray*}
\int  p(\theta \vert \tilde{ \mathcal{D} }) p(\tilde{ \mathcal{D} } \vert \mathcal{D} )  \dd \tilde{\mathcal{D}} = p(\theta \vert \lbrace (x^*_i, y_i) \rbrace_{i=1}^N),
\end{eqnarray*}
where $x^*_i = x_i + \argmax_{\Vert \gamma \Vert \leq \epsilon} L(\theta, x_i + \gamma, y_i) $. The MAP estimate of $\theta$ is  
\[
\theta^{MAP} = \argmax_{\theta} \left[  \log p(\theta \vert \lbrace (x^*_i, y_i) \rbrace_{i=1}^N) \right] 
= \argmin_{\theta} \left[ - \log \sum_{i=1}^N p(x^*_i, y_i \vert \theta ) - \log p(\theta)  \right], \]
which, according to \eqref{aux} is nothing else but the loss evaluated under the worst case transformation $x_i^*$.  Thus, AT is a special case of our workflow, in which we assume CK in the sense that the attacking model $p(\tilde{\mathcal{D}} \vert \mathcal{D})$ is deterministic; i.e., a degenerate distribution.


There are several ways of sampling from the conditionals \eqref{gibbs1} and \eqref{gibbs2}. \cite{math8111957} propose a scalable way of doing it leveraging efficient SG-MCMC sampling algorithms, and, in particular, stochastic gradient Langevin dynamics (SGLD, \citealp{welling2011bayesian}).
First, to account for the uncertainty that the defender has over the attacking model, the authors propose defining $p(\tilde{D}|D, \theta) \propto \exp{ \lbrace L(\theta, x_i, y_i) \rbrace }$.
%
Under SGLD, sampling iterations adopt the form
\begin{equation}\label{eq:bayes_attack}
x_{i, t+1} = x_{i, t} -\epsilon\,  \nabla_x (\log p(y|x_{i,t},\theta) + \log p(\theta))  + \xi_t,
\end{equation}
with $\xi_t \sim \mathcal{N}(0, 2\epsilon)$, and $t=1,\ldots,T$ with $x_{i,1} = x_i$. Note that this is a sampler from the distribution $p(\tilde{D}| D, \theta)$ and we approximate this distribution with the set of attacked samples $\lbrace x_{i}^* \rbrace_{i=1}^N$, setting $x_{i}^* = x_{i, T}$. Further uncertainties can also be accounted for; let us denote 
with  $\lambda$ the vector of hyperparameters of the optimizer, such as the step sizes $\epsilon_t$ or the number $T$ of iterations. Then, we have
$p(\tilde{D}| D, \theta) = \int p(\tilde{D} |D, \theta, \lambda) p (\lambda) d \lambda.$ To generate a perturbed data sample from the previous distribution, we need to sample from $p(\lambda)$. For instance, we could sample the step sizes $\epsilon_t$ from a beta distribution over $[ 1e-5, 1e-3 ]$, which are typical values in computer vision tasks using deep NNs; the number of iterations $T$ could 
be sampled from a Poisson distribution. Moreover, we could consider mixtures of different attackers, for instance by sampling a Bernoulli random variable and then choosing the gradient corresponding to either FGSM or another attack, such as \cite{7958570}'s. 

The attacker might have also uncertainty over the model the defender adopts, let it be the concrete model architecture or its parameters' values. Accounting for this would entail that the previous sampling scheme is done over an uncertain defender model $p(y|x, \theta)$ and start a hierarchy of level-$k$ thinking \citep{rios2012adversarial}, which can be computationally intractable. Instead, we propose mixing both steps and sample from the posterior distribution of the defended model rather than just arriving at $\theta^{MAP}$. To do so efficiently, if the model is optimized using gradient descent routines (as usual with deep NN models), we can again leverage SG-MCMC techniques to sample from the robust posterior $p(\theta \vert \lbrace (x^*_i, y_i) \rbrace_{i=1}^N)$, by repeating the following procedure for some number of training iterations:
\begin{enumerate}
    \item Sample perturbed samples $x_1, \ldots,  x_K \sim p(\tilde{D} | D, \theta)$ using the sampler from \eqref{eq:bayes_attack}, or from the natural distribution, for a mini-batch of size $K$.
    \item Update $\theta_{t+1} = \theta_{t} - \epsilon \nabla \sum_{i=1}^K L (\theta_{t}, x_i, y_i) + \mathcal{N}(0, 2\epsilon I)$ 
\end{enumerate}
In the end, we collect  $S$ samples $\lbrace \theta_i \rbrace_{i=1}^S$ from the robust posterior distribution. Then, given an instance $x$, we compute the predicted output $y_D$ approximating the posterior predictive utility using MC.

\section{Case studies} \label{sec:cases}
We illustrate the proposed defense mechanisms through the 
\textcolor{black}{motivating examples} from Section \ref{eg}.\footnote{Code to reproduce these experiments is available at \url{https://github.com/roinaveiro/aml_bayes}.} 

\noindent{\bf Case 1. Spam detection.}
Consider the set-up from Section 2. 

\textbf{Protection during operation.} 
For the first batch of experiments, \textcolor{black}{ we use the same algorithms in
Section \ref{eg}}. Recall the severe performance degradation \textcolor{black}{resulting from Good-Words-Insertion attacks} (Table 1, cols. 2 and 3).
Once the models are trained, we perform attacks
over the instances in the test set, solving 
problem \eqref{adv} for each test spam email, assuming certain values for the attacker's utilities and probability judgements.
%
%
%
%
$D$ is uncertain about the attacker's elements and models these uncertainties with random utilities and probabilities: 
we use beta distributions centered at the attacker's actual utility and probability values 
with variances chosen to guarantee that the distribution is concave in its support (they must be bounded from above by $\min \big \lbrace[\mu^2(1-\mu) ]/ (1+\mu), [\mu(1-\mu)^2] / (2-\mu) \big \rbrace$, where $\mu$ is the corresponding mean). The variance size informs about the degree of knowledge the defender is assumed to have about the attacker; 
reflecting a moderate lack of knowledge, we set the variance to be $10 \%$ of this upper bound. Of course, we are assuming certain degree of knowledge about the adversary, as the expected values of the random utilities and probabilities coincide with the actual values used by the attacker. We later study  how deviations from the assumed attacker behavior affect performance.

\textcolor{black}{ Having a model for the attacker, for each instance $x'$ of the test set,
the defender computes the robust adversarial posterior predictive distribution $\int p(y \vert x)p(x \vert x') \dd x$, and assigns $x'$ to the class maximizing the posterior predictive utility \eqref{def}. To compute the RAPPD, samples from $p(x' \vert x)$ are obtained leveraging the ability to sample from the attacker model and utilizing ABC as in \cite{gallego2020protecting}.} 


Column 4 in Table \ref{tab:cleanVSattack} compares the
average accuracy of the robustified during operation classifiers on tainted data.  
As can be seen, our approach allows us to reduce the performance degradation of the four original classifiers, showcasing the benefits of explicitly modeling the attacker's behavior in adversarial environments.
%
\textcolor{black}{  Interestingly, in the naïve Bayes case, 
our approach even outperforms the algorithm behaviour under untainted data} (column 2). This effect has been  observed also in \cite{naveiro2019adversarial} and \cite{goodfellow2014explaining} for other algorithms and application areas. This is likely due to the fact that the presence of an adversary has a regularizing effect, being able to improve the original accuracy of the base algorithm, \textcolor{black}{and} making it more robust.

The previous experiment used beta distributions centered around the values actually employed by the attacker to quantify the uncertainty about the attacker's utility and probability. It is natural 
to explore how deviations from the assumed values affect performance.
Our second batch of experiments tests the approach against an attacker whose utilities and probabilities are different from those 
assumed by the defender. In particular, for each attack, $A$ deviates uniformly around the assumed probability and utility. The size of the deviation is constrained to be less than 50\% the assumed value: if we center our beta distribution for, e.g., the attacker's probability at value $\mu$, the attacker will deviate from the assumed behavior in the range $(0.5 \cdot \mu, 1.5 \cdot \mu)$.
Thus, in this experiment, our beta distributions will be centered around 
{\em wrong values}. 
We set the variance of the beta priors to be relatively high, at $50 \%$ of the upper bound, and compare our approach with the CK one, in which the elements of the attacker are assumed to be known, and thus are point masses (on wrong values). 

Table \ref{tab:acra_vs_ck} shows average accuracy plus minus one standard deviation (estimated through repeated hold out validation) of the four algorithms on attacked data without defense (col. 2), the standard 
 CK defense (col. 3)
and, finally, our ARA defense (col. 4). Note first  the overall performance drop with respect to the results in Table \ref{tab:cleanVSattack} col. 4:
when the attacker deviates from his assumed behavior, the performance of both ARA-based and CK defenses is lower. However, we can also observe that the ARA-based defense outperforms the CK \textcolor{black}{defense} for all classifiers: when the attacker deviates from the assumed behavior, accounting for the uncertainty over his elements is beneficial. This experiment showcases the increase in robustness due to modelling uncertainty in scenarios in which CK is not realistic.
%
\begingroup
\renewcommand{\arraystretch}{0.5}
\begin{table}[!h]
\caption{Average accuracy plus minus one standard deviation
	of four algorithms on attacked data without defense (col 2);
	 with CK defense (col 3); and with ARA defense (col 4).}
	\centering
    {\small
	\begin{tabular}{ccccccc}
		\toprule
		\textbf{Classifier} &  \textbf{Acc. Taint.} & \textbf{Acc. CK Taint.} & \textbf{Acc. ARA Taint.}\\
		\midrule
		Naive Bayes   & $0.793 \pm 0.005$ & $0.867 \pm 0.004$ & $0.883 \pm 0.005$        \\
		Logistic Reg.  & $0.687 \pm 0.008$    & $0.803 \pm 0.007$  & $0.864 \pm 0.005$     \\  
		Neural Network  &      $0.774 \pm 0.007$     &     $0.767 \pm 0.007$  & $0.792 \pm 0.006$         \\
		Random Forests   &    $0.682 \pm 0.005$  &    $0.819 \pm 0.007$     & $0.821 \pm 0.007$       \\
		\bottomrule
	\end{tabular}%
	}
	\label{tab:acra_vs_ck}%
\end{table}
\endgroup

\textbf{Protection during training.}\label{sec:exp_scalable}
We next assess ARA based  robustification during training.
This requires the underlying model
to be differentiable in the parameters, thus
leaving just two candidates among the original models: 
logistic regression and NN. 
Both models can be trained using SGD plus noise methods to obtain uncertainty estimates from the posterior.
Next, we attack the clean test set using the procedure 
in Section \ref{eg} and evaluate the performance of 
our robustification proposal. 
 Since we are dealing with discrete attacks, we cannot use the uncertainty over attacks as in (\ref{eq:bayes_attack}). Instead, we model it using the distribution $p(x'|x)$ and take samples from it as discussed in 
step 3 from our Section \ref{KK} workflow. 
We evaluate the Bayesian predictive distribution using $S=5$ posterior samples obtained after $T=2000$ SGLD iterations, and present the results in Table \ref{tab:cleanVSattack}, col. 5.
%
Observe again that the proposed robustification process protects differentiable classifiers, 
recovering from the degraded performance under attacked
data. Note that the robustified algorithms achieve even higher accuracies than those attained by the original classifier over clean data, due to the regularizing effect mentioned above.
\hfill $\triangle$
\vspace{0.1in}

\begin{figure}[h]
     \centering
     \begin{subfigure}[b]{0.49\textwidth}
         \centering
         \includegraphics[width=\textwidth]{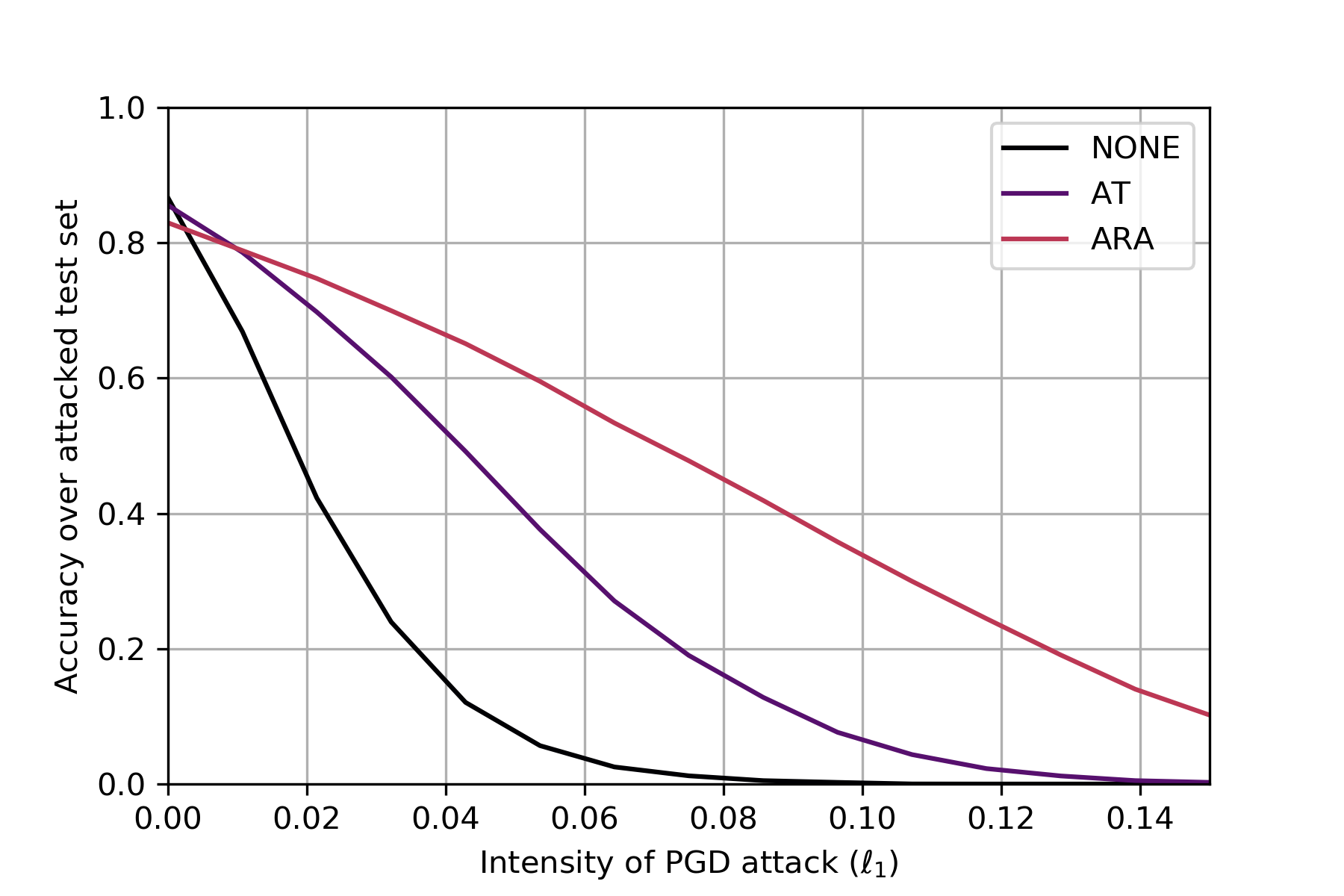}
         \caption{Fashion-MNIST dataset.}
     \end{subfigure}
     \hfill
     \begin{subfigure}[b]{0.49\textwidth}
         \centering
         \includegraphics[width=\textwidth]{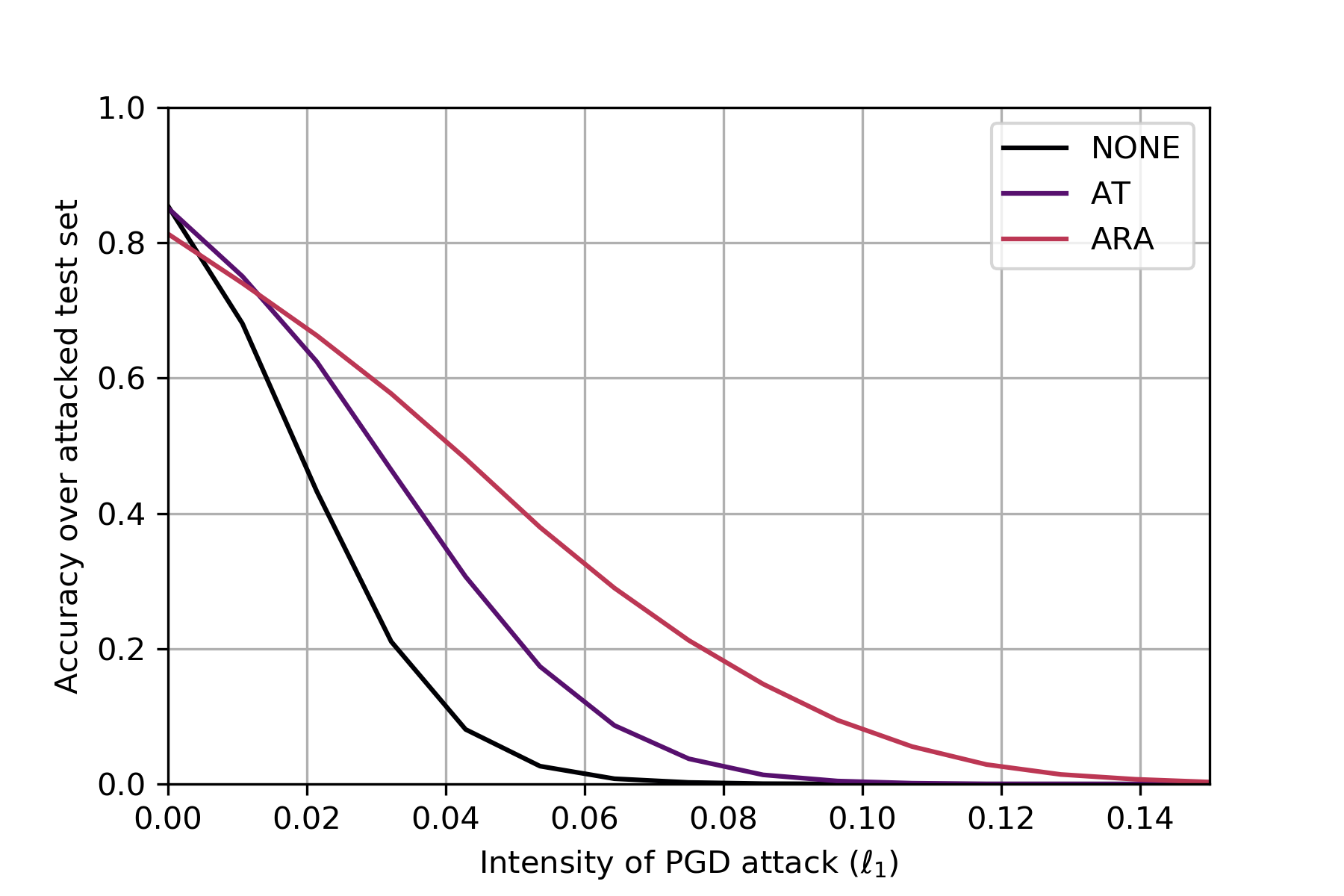}
         \caption{Kuzushiji-MNIST dataset.}
     \end{subfigure}
        \caption{Robustness of a deep network against the PGD attack under three defense mechanisms 
(NONE, AT, ARA). (a) depicts the security evaluation curves for the attacked Fashion-M. dataset. (b) depicts the respective curves for the attacked Kuzushiji-M. dataset.}
       \label{fig:comparison}
\end{figure}

\noindent{\bf Case 2. Vision.} 
When the input data is high-dimensional (such as with images), our ARA robustification at operation easily becomes computational intractable.
We thus robustify model at training. 
For illustration purposes,
we perform additional experiments using two
benchmarks: Fashion-MNIST, a clothing classification problem \citep{xiao2017fashion}, and Kuzushiji-MNIST, a traditional Japanese handwritten character recognition problem \citep{clanuwat2018deep}. For both datasets we trained standard deep NNs over their respective training sets (consisting of 60.000 images each) using SGD (i.e., no defense), adversarial training (AT defense), and our robustification procedure from Section \ref{KK} (ARA defense). We then attacked the respective test sets using 5 iterations of PGD, with varying attack intensities (the step-size $\epsilon$ in Eq. (\ref{eq:bayes_attack})), and evaluated the accuracies of both models under these attacked test sets. Figure \ref{fig:comparison} displays these results. Note that our scalable approach from Section 4 offers fairly superior robustification defenses compared  
to the AT defense mechanism, showing that \textcolor{black}{ incorporating
the uncertainties provided by the ARA methodology }
has additional benefits when adversarially training a ML model in diverse datasets. \hfill $\triangle$

\section{Conclusion}
We have provided a review of key approaches, models, and concepts in AML. This area is of major importance in security and cybersecurity
to protect systems that increasingly rely on
ML algorithms \citep{COMITER2019, trend2020}. The pioneering work by \cite{adversarialClassification2004}  
framed most of this research within the game theory realm, with entailed CK conditions which hardly hold in AML security contexts.

\textcolor{black}{We have proposed a Bayesian alternative to AML. Its 
 main difference with respect to previous approaches is that unrealistic CK conditions are not entailed. As a consequence of the increased realism, the resulting models are more robust, especially with respect to deviations in the assumed attacker behaviour,  given the 
 better reflection of the involved uncertainties as empirically illustrated. Our Bayesian framework enjoys greater flexibility than previous game-theoretic approaches, some of which can be framed as limit or degenerate cases of our proposal. However, the increased robustness and flexibility comes at a higher computational and modelling costs. Investigating how to alleviate these costs is an open research question. Related to this, we sketch several promising avenues for future work.}

A promising research line consists of 
developing efficient algorithms for approximate Bayesian inference with robustness guarantees. For example, regarding opponent modeling in sequential decision making, an agent has uncertainty over her opponent type initially; as more information is gathered, she might reduce her uncertainty via Bayesian updating. 
Similarly, work in robust Bayesian analysis 
\citep{rios2000robust}, in particular referring to likelihood robustness, is relevant.
Not taking into account an attacked
data generation process is an example of model misspecification; robustness of Bayesian inference to such issue has been revisited recently in  \cite{miller2019robust}. 

There are also several enhancements aimed at improving operational aspects of the framework.
For example, we discussed only problems with two agents.
It would be relevant to deal with multiple agents, including cases in which agents on \textcolor{black}{attack or defense attempt to} cooperate. 
There is also  potential in new algorithmic approaches.
Exploring gradient-based techniques for bi-level optimization problems arising in AML is a fruitful line  \citep{naveiro2019gradient}. More efficient MCMC samplers \textcolor{black}{can} be adopted in the proposed workflow \citep{gallego2018sgmcmc}. 
Recall that our framework essentially goes through simulating from the attacker problem to forecast attacks and then optimizing for the defender to find her optimal decision. This may be computationally demanding and we could explore single stage approaches, such as augmented probability simulation \citep{ekin2022augmented}.

As mentioned in Section \ref{sec:review}, adversarial versions of various ML problems has been studied. However, further research is required in unsupervised learning, including clustering methods,
dynamic linear models \citep{naveiro2021adversarial}, natural language
processing models \citep{wang2019knowing}, and in RL, including policy gradient \citep{10.5555/3172077.3172414} and extensions to semi-Markov Decision Processes \citep{du2020modelbased}.

Applications, such as those presented in \cite{COMITER2019} and \cite{trend2020}, are abound. We mention four of direct interest to us: (i.) the development of defenses against fake news;
(ii.) the development of robust ADS algorithms \citep{caballero2021decision};
(iiii.) the use of AML for improving counterfactual inference in observational studies \citep{johansson2016learning}; and (iv.) leveraging ideas from causal inference to improve adversarial robustness \citep{scholkopf2021toward}. Several recent works, for instance, aim to improve the robustness of deep NNs for image classification by leveraging a causally informed model of unseen perturbations or adversarial examples \citep{NEURIPS2020_02ed8122,zhang2021adversarial}.

\section*{Ackowledgements}
The authors acknowledge the support of the National Science Foundation under Grant DMS-1638521 to the Statistical and Applied Mathematical Science Institute (SAMSI), NC, USA. RN acknowledges the support of CUNEF University. DRI is supported by the AXA-ICMAT Chair and the Spanish Ministry of Science program PID2021-124662OB-I00. This work supported by Severo Ochoa Excellence Programme CEX-2019-000904-S,
  the European Union's Horizon 2020 Research and Innovation Programme under Grant Agreement No. 101021797 (Starlight) and 815003 (Trustonomy), the US NSF grant DMS-1638521 and grants from the FBBVA (Amalfi), EOARD (FA8655-21-1-7042) and AFOSR (FA-9550-21- 1-0239). VG acknowledges support from grant FPU16-05034.
\bibliographystyle{asa}
\begin{singlespace}
\small{
\bibliography{references}
}
\end{singlespace}

\end{document}